\documentclass[a4paper]{article}
\usepackage[a4paper,top=3cm,bottom=2cm,left=3cm,right=3cm,marginparwidth=1.75cm]{geometry}

\usepackage[english]{babel}
\usepackage[utf8x]{inputenc}
\usepackage[T1]{fontenc}
\usepackage{soul}

\usepackage{lipsum}
\let\OLDthebibliography\thebibliography
\renewcommand\thebibliography[1]{
	\OLDthebibliography{#1}
	\setlength{\parskip}{0pt}
	\setlength{\itemsep}{0pt plus 0.3ex}
}

\usepackage{amsthm}
\usepackage{amsmath}
\usepackage{amssymb}
\usepackage{tikz-cd}
\usepackage{multirow}
\usepackage{comment}
\usepackage{algpseudocode, algorithm}

\newtheorem{Theorem}{Theorem}[section]
\newtheorem{Proposition}[Theorem]{Proposition}

\title{Stable Topological Summaries for Analyzing the Organization of Cells in a Packed Tissue}
\author{Nieves Atienza, Maria-Jose Jimenez, Manuel Soriano-Trigueros}

\date{}

\graphicspath{{figures/}}

\begin{document}
	\maketitle
	\begin{center}
		\centering Department of Applied Mathematics I, University of Seville\\
		\{natienza, majiro, msoriano4\}@us.es
	\end{center}
\begin{abstract}
	We use Topological Data Analysis tools for studying the inner organization of cells in segmented images of epithelial tissues. More specifically, for each segmented image, we compute different persistence barcodes, which codify lifetime of homology classes (persistent homology) along different filtrations (increasing nested sequences of simplicial complexes) that are built from the regions representing the cells in the tissue. We use a complete and well-grounded set of numerical variables over those persistence barcodes, also known as topological summaries. A novel combination of normalization methods 
	for both, the set of input segmented images and the produced barcodes, allows to prove stability results for those variables with respect to small changes in the input, as well as invariance to image scale. Our study provides new insights to this problem, such as a possible novel indicator for the development of the drosophila wing disc tissue or the importance of centroids distribution to differentiate some tissues from their CVT-path counterpart (a mathematical model of epithelia based on Voronoi diagrams).  We also show how the use of topological summaries may improve the classification accuracy of epithelial images using Random Forests algorithm.
\end{abstract}

\section{Introduction}

Epithelia  are  packed  tissues  formed  by  tightly  assembled  cells, with almost no intercellular spaces. 
There are many studies in the literature  focused on their natural organization, 
since changes on it may indicate an early onset of disease \cite{cell1,cell2}. 
Therefore, being able to quantify differences in the spatial distribution of cells is an interesting problem.\\
In some of these tissues,
their apical surfaces are similar to convex polygons forming a natural tessellation.
That allows to identify each cell with a polygon with as many sides as neighboring cells. 
The study of epithelial organization has been mainly focused on the polygon distributions, 
that is, 
the distribution of the number of neighbors (sides) of the cells (polygons). 
In \cite{cell3}, 
the authors looked for differences in polygon distributions on two proliferative stages of drosophila wing disc. 
In later studies, 
the concept of centroidal Voronoi tessellation (CVT) was used, 
which is a Voronoi diagram where the point generating each region coincides with its centroid. 
The Lloyd algorithm is an iterative algorithm that, 
starting from a random cloud of points, 
produces a series of Voronoi diagrams, 
that we will denote by ($CVT_1,\,CVT_2,$ $\,CVT_3,\ldots)$, 
that converge to a CVT \cite{convergencia}. 
Such a sequence of Voronoi diagrams is called CVT-path in \cite{embo}. 
There, 
the authors compared the polygon distributions of images of natural packed tissues with those of the CVT-path and showed that the former fit to certain distributions given inside the CVT-path.
A different approach was developed in \cite{epigraph}, 
were the authors provided an image analysis tool implemented in the open-access platform FIJI, 
to quantify epithelial organization based in computational geometry and graph theory concepts. 
More specifically, 
considering the {\em contact graph}, 
that is, 
the graph generated by the cells (vertices) and the cell-to-cell contacts (edges), 
they searched {\em locally} for specific motifs represented by small subgraphs (graphlets) to characterise the tissue.\\
The previous approaches have two main problems:
assuming that the cells are similar to convex polygons and that the analysis is mostly related to local features ignoring other aspects of the contact graph, as for example, what types of polygons are connected among them.\\
As pointed out by P. Villoutreix in his thesis,
the standard topological analysis of  complex networks is very limited in this context,
since all contact graphs are planar (see \cite{Villoutreix}[Sec. 8.2.2]).\\
In \cite{Villoutreix},
the author proposed the use of topological data analysis (TDA) as a possible solution to obtain richer information than just the polygon distribution while keeping the stability of the method respect to the inner organization of the tissue.\\
In this paper we prove these two postulates.
In particular,
we will formalize the intuitive notion of inner organization of the tissue using its contact graph and its spatial centroid distribution. 
In addition, we will show that TDA may work when cells are not convex-like. 
Finally,
we add TDA variables to a machine learning workflow to improve the classification of images coming from cellular tissues.
We also explain some interpretations of the TDA variables at the tissue level, 
giving new insights about the organization of the cell.
\subsection{Previous topological data analysis approaches}

Recall that topology is the branch of mathematics that deals with properties of space that keep invariant under continuous transformations. 
These properties may be extremely important when the space is a network.\\
Nowadays, 
TDA is spreading as a useful approach in very different scientific fields, 
playing an increasing role in biological and, 
in general, 
biomedical imaging.  
Its main analysis tool, {\em persistent homology} \cite{ELZ00,ZC05}, has been successfully applied in solving problems such as tumour segmentation \cite{tumor}, 
analysis of biological networks \cite{inmuno} or diagnostic of chronic obstructive pulmonary disease \cite{lung}. 
Persistent homology studies the evolution of {\em homology classes} and their life-times (persistence) in an increasing nested sequence of spaces (called a {\em filtration}). A filtration could be thought as a multiscale combinatorial model that represents topological (and somehow, geometric) information of the data.  The result of persistent homology computation can be codified as a combinatorial invariant, called the {\em persistence barcode}, which plays as a topological signature of the filtration, and therefore of the original data set.\\
In \cite{Villoutreix},
a model for analyzing the contact graph of the cell tissues using {\em \textbf{sub}} and {\em \textbf{sup} filtrations} (see Section~\ref{sec:clique}).
To our knowledge,
this was the first experiment relating epithelial tissues with TDA.
Its nature was exploratory and the results similar to the ones obtained using the mean and variance of the degree of the cells.
Another problem of this analysis was that the polynomial variables used for analyzing the barcodes were not stable with respect to the bottleneck distance (see Section~\ref{sec:stability}).
Defining stable polynomial variables was one of the main reasons why tropical coordinates were defined in \cite{tropical}.
We have found the work in \cite{Villoutreix} extremely useful as a first approach and this paper can be seen as a continuation of it.
Independently,
a similar approach was presented in the conference paper \cite{IWCIA17}.
In this case,
persistent entropy was used instead of polynomial variables.
Again,
stability was not guaranteed,
since the number of bars of each barcode was not fixed, as  
required in \cite{stability} for a stable result. 
Finally,
another approach was introduced in the  conference paper \cite{ctic}.
In that case, instead of using the contact graph of the cells,
 spatial distribution of their centroids was studied,
using the {\em \textbf{alpha} filtration} (or alpha-complex) which was constructed over the {\em Delaunay complex} generated by the set of centroids  \cite{computational}.
Nevertheless,
keeping the infinity bar in the barcodes made the summaries depend on the original scale of the image,
introducing bias in the analysis.
One of the major motivations of this paper was to solve that problem.
Needless to say that finding a setting for normalizing the barcodes (so that they can be compared), while guaranteeing the stability of the variables used to analyze them is far away from being trivial and require a mathematical analysis even for a specific case study like this one.\\
Recently, a paper applying TDA to cell images has been published \cite{collective}. Their approach uses cycles to analyze the clustering of epithelial cells as self-propelled particles (not forming packed tissues).
\subsection{Overview of the paper}

In Section~\ref{sec:2},
we will modify the methods appearing in \cite{Villoutreix} and \cite{ctic} to guarantee the stability in all the procedure. 
Mathematical proofs to support correctness
are provided.
No assumption about convexity of the cells is needed.
In Section~\ref{sec:3},
a rigorous statistical analysis of the results for tissues (both, epithelium and CVTs) is carried out using TDA variables.
Also,
these variables will be used to improve the performance of a Random Forests classification of the tissues based on their neighbors distribution.
Interpretation of some of the variables at the tissue level are provided.
Finally,
we will summarize the results in Section~\ref{sec:4} and propose new interesting questions arising from this paper that might be of interest for different fields, such as developmental biology,
pattern recognition or TDA.
\section{Materials and Methods}\label{sec:2}

Our aim is to assign to each epithelial image an invariant, called persistence barcode, representing inner topological and geometrical information. We would like to analyze these persistence barcodes using numerical variables. There are three main difficulties:
\begin{itemize}
 	\item Finding a correct data normalization which does not include bias in our analysis due to the number of cells or the scale of the image;
    \item Guaranteeing our variables only measure topological-geometrical properties. In particular, they should be invariant to rotation or scale changes.
    \item Proving  robustness  of  our  variables with respect to the cell organization. 
\end{itemize}
\subsection{Input data}

Our method is suitable for the topologial analysis of the organization of segmented regions that partition a portion of plane. 
In this paper, 
the images for our experiments come from several types of (real) epithelial tissues as well as different mathematical tessellations. 
The segmented images of epithelial tissues considered are available as suplementary material in the article \cite{images}. 
More specifically, 
they are images of chicken tissues: chicken embryonic ectoderm (cEE) images, chicken neural tube (cNT) images; 
and Drosophila tissues: Drosophila notum prepupa (dNP) images, wing disc in the larva (dWL) and prepupal (dWP) stages of development.
The tissues dWL and dWP are taken from two proliferative stages separated by only 24 hours of development, 
so they are really hard to distinguish.
Further information about the way these images were obtained and segmented can be found in \cite{images}.\\
Besides, 
since the morphology of cells in some epithelial tissues is commonly approximated by Voronoi tessellations \cite{KJRS2016}, 
we also consider the so-called CVT-path: 
take a random set of points on the plane and, 
iteratively, 
compute the Voronoi diagram $CVT_i$ from them and its set of centroids as seed points for next iteration ($i+1$).
Information of how the CVT-path is generated can be found in \cite{embo}.\\
In order to obtain information from the regions like the centroids,
the \emph{Matlab} funtion {\em regionprops} was used. 
For the contact graph,
a small dilatation was performed on each region and labels of adjacent regions reached by the dilation were retrieved.
Cells are said to be {\em valid} if they do not touch the exterior limits of the image. Only valid cells will be processed.
The data extraction procedure, together with the whole code, can be found in 
a publicly available repository.\footnote{github.com/Cimagroup/topo-summaries-for-packed-tissues}
\subsection{Normalization and cell selection}\label{sec:number}

\begin{figure}
    \centering
        \begin{tabular}{c c c}
            \includegraphics[]{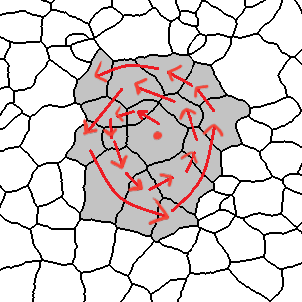} &\qquad\qquad& \includegraphics[scale = 0.32]{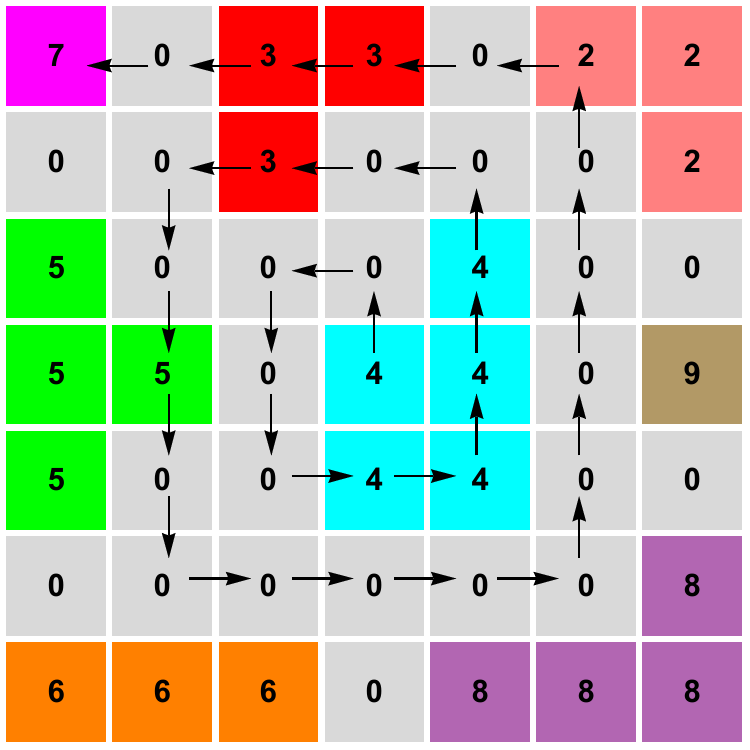}
        \end{tabular}
    \caption{Left: 
    Intuition behind the algorithm.
    We pick cells following a spiral until the desired number is reached.
    Right: 
    Toy example. 
    Taking as input $N=5$,
    the output is the set of labels $\mathcal{C}=\{4,3,5,2,7\}$. Boundary pixels are labelled by $0$. }
    \label{fig:spiral}
\end{figure}

In order to avoid the bias induced by the number of valid cells in each tissue,
we will consider always the same number of cells from each image to proceed to the topological analysis. 
Besides, 
this normalization is key to prove some stability results in Subsections~\ref{sec:normalization} and \ref{sec:variables}. 
Although
a higher number of cells is, 
in general, 
better to have a more global picture of the organization, 
the amount considered will be constrained by the minimum number of cells in the images of the given database.  
The number of valid cells
in the whole set of images range from $140$ to $1102$,
see Table~\ref{tab:number}.
Then, 
as a starting point, 
we can fix $N=140$ as the number of cells picked. 
Unfortunately,
a rare event happens in the first cEE image: some valid cells are completely surrounded by non-valid cells making them disconnected from the rest.
We have decided to dismiss it as an outlier (representing a $2.8\%$ of the total sample).
Then,
we will fix $N=187$, as it is the second minimun number of cells appearing in Table~\ref{tab:number}.
We follow Algorithm~\ref{alg:spiral} \cite{ctic} to select the desired number of cells in each image.
In Figure~\ref{fig:spiral} an intuitive idea of how the algorithm works is given.
From now on, 
we will denote as valid cells only to the ones selected by this algorithm.

\begin{table}
\centering
\caption{The number of valid cells in each image of the epithelial tissues.}\label{tab:number}
\vspace{0.3cm}
\footnotesize
\begin{tabular}{c|c|c|c|c|c|c|c|c|c|c|c|c|c|c|c|c|}
             & \textbf{1} & \textbf{2} & \textbf{3} & \textbf{4} & \textbf{5} & \textbf{6} & \textbf{7} & \textbf{8} & \textbf{9} & \textbf{10} & \textbf{11} & \textbf{12} & \textbf{13} & \textbf{14} & \textbf{15} & \textbf{16} \\ \hline
\textbf{cEE} & 140        & 206        & 229        & 241        & 385        & 380        & 261        & 187        & 405        & 246         & 327         & 204         & 348         & 270         &             &             \\
\textbf{cNT} & 666        & 661        & 566        & 574        & 669        & 532        & 420        & 592        & 744        & 527         & 594         & 473         & 704         & 748         & 469         & 834         \\
\textbf{dNP} & 513        & 723        & 588        & 525        & 439        & 823        & 1102       & 533        & 309        & 575         & 302         & 375        &             &             &             &             \\
\textbf{dWL} & 432        & 556        & 485        & 525        & 501        & 936        & 890        & 790        & 977        & 913         & 606         & 835         & 785         & 748         & 622         &             \\
\textbf{dWP} & 748        & 806        & 566        & 415        & 454        & 654        & 752        & 713        & 504        & 430         & 387         & 516         & 419         & 455         & 277         & 257  \\ \hline      
\end{tabular}
\end{table}
\begin{algorithm}[ht]
    \caption{Spiral selection of regions}\label{alg:spiral}
    \begin{algorithmic}[1]
    \Procedure{SPIRAL}{$M,n$} \Comment{$M$ is an image and $n$ a number}
    \State $\mathcal{C} :=\{\,\}$
    \State $(x,y) := center\left( M \right)$ \Comment{central coordinates of $M$}
    \If {$M(x,y) \neq 0$}
    	\State $\mathcal{C}:= \{M(x,y)\}$
    \EndIf 
    \State $i := 0$
    \While{$\#\mathcal{C} < n$} \Comment{$\#$ is the number of elements}
    	\State $i := i+1$
    	\For {$j \in (1, \ldots, i)$} \Comment repeat $i$ times
    		\If {$\#\mathcal{C} < n$} 
    		\State $x := x + (-1)^{i}$
    			\If {$M(x,y) \neq 0$ and  $M(x,y) \notin \mathcal{C}$}
    				\State 	$\mathcal{C}:=\mathcal{C}\cup \{M(x,y)\}$
    			\EndIf
    		\EndIf
    	\EndFor
    	\For {$j \in (1, \ldots, i)$} \Comment repeat $i$ times
    		\If {$\#\mathcal{C} < n$}
    			\State $y := y + (-1)^{i}$
    			\If {$M(x,y) \neq 0$ and  $M(x,y) \notin \mathcal{C}$}
    				\State 	$\mathcal{C}:=\mathcal{C}\cup \{M(x,y)\}$
    			\EndIf
    		\EndIf
    	\EndFor			
    \EndWhile
    \State \textbf{return} $\mathcal{C}$ \Comment{return the first $n$ labels around the center}
    \EndProcedure
    \end{algorithmic}
\end{algorithm}
\subsection{Simplicial complexes and filtrations}

A \emph{$k$-simplex} (or simplex of dimension $k$) in $\mathbb{R}^m$ is the convex hull of a set of $k+1$ affinely independent points $ \tau = \{ p_0, \ldots, p_k \}$. The points of $\tau$ are called the \emph{vertices} of $\tau$ and the subsets of $\tau$ form the \emph{faces} of $\tau$. 
That is, each $\ell$-simplex contained in $\tau$ with $0\leq \ell<k$ is called a \emph{face} of $\tau$.
A (geometric) \emph{simplicial complex} $\mathcal{K}$ is formed by a set of simplices satisfying:
\begin{enumerate}
	\item Every face of a simplex in $\mathcal{K}$ is also in $\mathcal{K}$.
	\item The intersection of any two simplices in $\mathcal{K}$ is either a face of both simplices or the empty set.
\end{enumerate}
The \emph{dimension} of a simplicial complex is the maximum of the dimensions of its simplices.
The combinatorial description of $\mathcal{K}$ as finite subsets of the whole set of vertices $V$ (without considering the geometric embedding in $\mathbb{R}^m$) is known as an abstract simplicial complex. In the following, 
when we refer to a simplicial complex we mean an abstract simplicial complex.
\\
A \emph{filtration} over a simplicial complex $\mathcal{K}$ is a finite nested sequence of simplicial subcomplexes

\[
\mathcal{K}_{1} \subset \mathcal{K}_{2} \ldots \subset \mathcal{K}_{r} = \mathcal{K}
\]
It is commonly defined using a monotonic function $f: \mathcal{K} \rightarrow \mathbb{R}$ i.e. for any two simplices $\delta,
\tau \in \mathcal{K}$,
if $\sigma$ is a face of $\tau$, 
then $f(\sigma) \leq f(\tau)$. 
That way, 
if $a_1 \leq \ldots \leq a_r$ are the function values of all the simplices in $\mathcal{K}$,   
then the subcomplexes $\mathcal{K}_i = f^{-1}(-\infty, a_i]$, 
for $i=1\ldots r$ define a filtration over $\mathcal{K}$.
We may call $f$ a filtration when we actually refer to the filtration induced by $f$.\\
We will use three types of filtrations:
the clique complex filtration \textbf{sub},
the clique complex filtration \textbf{sup},
and the Vietoris-Rips filtration \textbf{rips}.
\subsubsection{The \textbf{sub} and \textbf{sup} filtrations}\label{sec:clique}

\begin{figure}
	\centering
	\includegraphics[width=\textwidth]{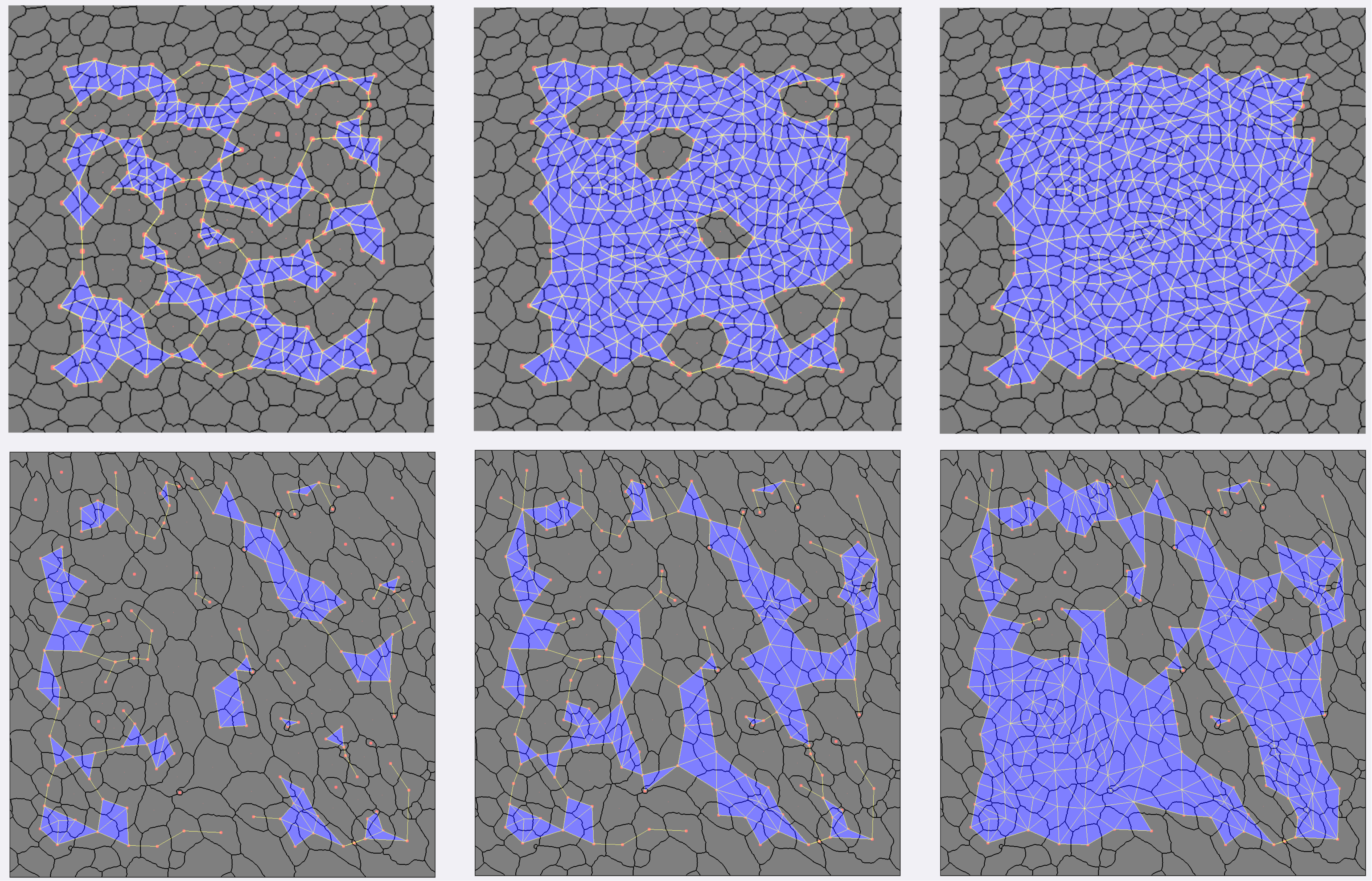}
	\caption{An example of the sub filtration for $i=6,\,7,\,8$. The top row corresponds to dNP and the bottom to cEE.}
	\label{fig:filt_sub_dnp_cee}
\end{figure}

Since we know which valid cells are neighbors between them,
we can build a graph representing this relation as edges.
We denote it as a \emph{contact graph}.
We construct the \emph{clique complex} of a graph, 
$CK$, 
adding a $k$-simplex $[x_0, \ldots, x_{k}]$ whenever the graph has a clique formed by the vertices $x_0, \ldots, x_{k}$.
Define the \textbf{sub, sup} filtration \cite{Villoutreix} over the clique complex of a contact graph using the following functions,

\begin{align*}
f_{sub}(\sigma) &= \max\{ VN(x) : x \in \sigma \}; \\
f_{sup}(\sigma) &= \max\{ 15 - VN(x) : x \in \sigma \};
\end{align*}
where $VN(x)$ is the number of valid neighbors of the cell (i.e. the degree of the vertex) $x$ and $\sigma$ a face of the simplicial complex.
We use the value $15$ in the \textbf{sup} filtration since it is rare to find a cell with such a number of neighbors, 
and in fact, there is not such a cell in our samples.
Note that both \textbf{sup} and \textbf{sub} only carry information about the topology of the contact graph of the cells 
(it is a topological invariant).
See Figure~\ref{fig:filt_sub_dnp_cee} for two examples.
\subsubsection{The \textbf{rips} filtration}\label{sec:rips}

\begin{figure}
	\centering
	\includegraphics[scale=0.6]{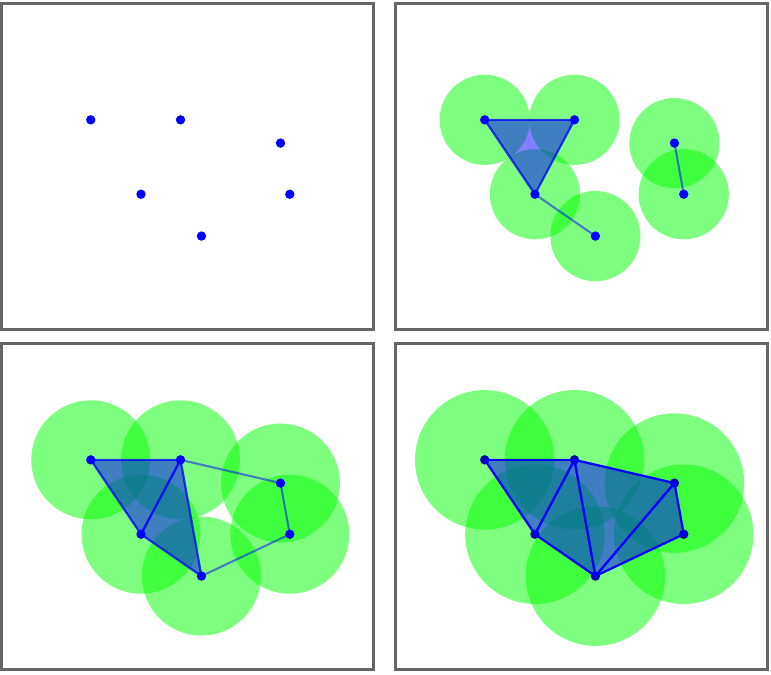}
	\caption{ An example of a \textbf{rips} filtration with $6$ points in the Euclidean plane.
	Note that a simplex arises when the distance between the corresponding centroids is smaller than or equal to twice the radius. 
	}
	\label{fig:rips}
\end{figure}

Another strategy is to obtain the centroid of each cell and study their distribution.
In order to do that,
we will use the Vietoris-Rips filtration \cite{computational}.
It is constructed using the function

\[
	f_{rips}(\sigma) = \max_{p,q \in \sigma} d(p, q),
\]
Over the simplex generated from the whole set of
centroids. 
See Figure~\ref{fig:rips} for an example.
It is important to emphasize that the immersion of the point clouds in the image to $\mathbb{R}^2$ depends on the ``distance per pixel relation'' of the original image,
and so does \textbf{rips}. Then, \textbf{rips} is not scale-invariant.
We would like to eliminate this bias with a normalization process, 
but due to stability arguments we will apply it in the next step and not directly on the filtration.
\subsection{Persistent homology and barcodes}

\begin{figure}[h!]
	\begin{center}
		\begin{tabular}{ |c|| c| }
			\hline\multicolumn{2}{|c|}
			{\hspace{0.1cm}
				\includegraphics [scale=0.4] {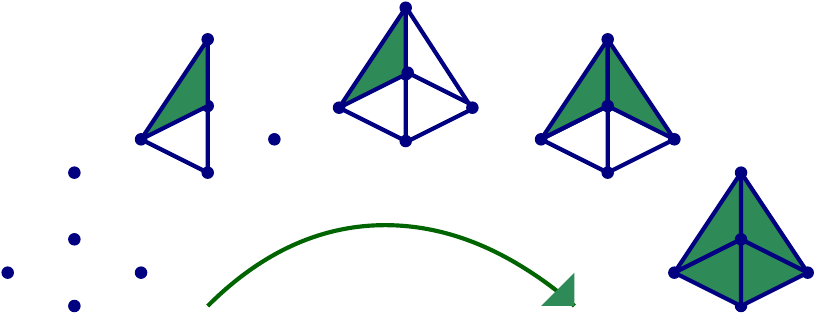}
				\hspace{0.1cm}}
			\\
			\hline
			\hline
			\begin{minipage}{.32\textwidth}
				\vspace{0.3 cm}
				\includegraphics [scale=0.25] {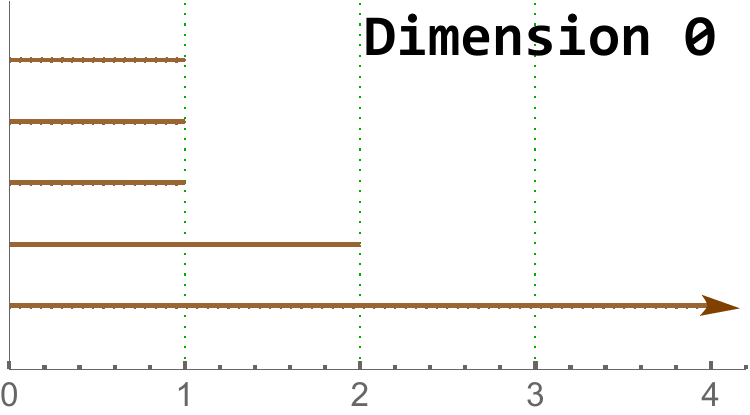}
			\end{minipage}
			&
			\begin{minipage}{.32\textwidth}
				\vspace{0.3 cm}
				\includegraphics [scale=0.25] {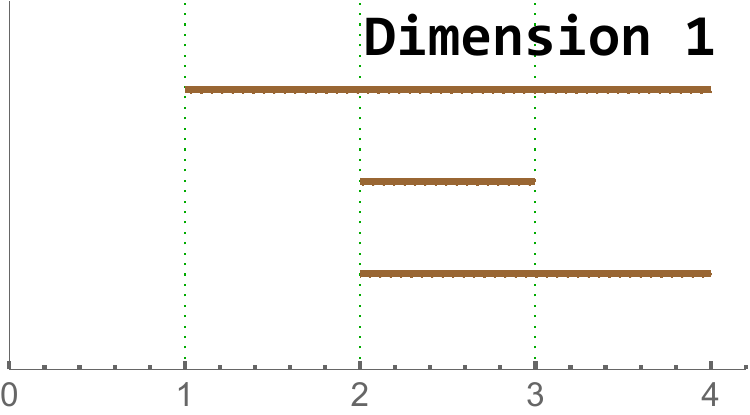}
			\end{minipage}
			\\
			\hline
		\end{tabular}
	\end{center}
	\caption{ Top.- example of a filtration $\mathcal{K}$. Bottom.- barcodes representing connected components and cycles. Note there is a $[0, \infty)$ bar in the $0$ dimensional persistent homology. In this example: $\beta_{2,2}^1 = 3,\, \beta_{1,2}^1 = 1,\, \beta_{2,3}^1 = 2,\, \beta_{1,3}^1 = 1$ and then $[2,3)$ appears once in the $1$-dimensional barcode}\label{fig:barcode_example}
\end{figure}

Intuitively,
homology formalize the notion of $m$-dimensional hole.
A $0$-dimensional hole is a connected component, 
a $1$-dimensional hole is a tunnel (or a cycle in a graph), 
a $2$-dimensional hole is a cavity, and so on.
More specifically, 
 homology gives a procedure to assign to a simplicial complex $\mathcal{K}$, a vector space $H_m(\mathcal{K})$ as follows:\\
First,
define the $m$-chains of $\mathcal{K}$, 
$C_m(\mathcal{K})$, 
as the vector space over a field $\mathbb{F}$, with basis the set of $m$-dimensional simplices of $\mathcal{K}$.
In this paper,
we will use $\mathbb{F} = \mathbb{Z}_2$.
If $\tau$ is an $m$-simplex,
define the boundary operator in each $m$-simplex as $\partial_m(\tau) = \sum_{\sigma \in F} \sigma$ where $F$ are the $n-1$ faces of $\tau$.
Then,
extend it linearly to obtain $\partial_m : C_m(\mathcal{K}) \rightarrow C_{m-1}(\mathcal{K})$.
Homology is the vector space

\[ H_m(\mathcal{K}) = \dfrac{\ker \partial_{m}}{\emph{im } \partial_{m+1}}. \]
where $\ker$ is the Kernel of $\partial$ and \emph{im } is the Image.
Each of the classes of $H_m(\mathcal{K})$ can be seen as a hole of $\mathcal{K}$.
The $m$-betti number, 
$\beta_m = \dim H_m(\mathcal{K})$,
is interpreted as its amount of $m$-dimensional holes.\\
Besides, 
if we have two simplicial complexes $\mathcal{K}_a \subset \mathcal{K}_b$,
homology induces a linear map $f_{a,b}$ between $H_m(\mathcal{K}_a)$ and $H_m(\mathcal{K}_b)$. 
In this case, 
$\beta^m_{a,b} = \dim f_{a,b} H_m(\mathcal{K}_a)$ can be seen as the number of $m$-dimensional holes shared by both simplicial complexes.\\
Persistent homology study how the $m$-dimensional holes appear and disappear in a filtration $\mathcal{K}_{a_1} \subset \mathcal{K}_{a_2} \ldots \subset \mathcal{K}_{a_r}=\mathcal{K}$.
Fix a pair of numbers $i = 1, \ldots, r$ and $j =2, \ldots r+1$. Following the previous reasoning,
note that the value

\[
c_{a_i,a_j}^m = \beta_{a_i,a_{j-1}}^m - \beta_{a_{i-1},a_{j-1}}^m - \beta_{a_i,a_{j}}^m + \beta_{a_{i-1},a_j}^m
\]
can be interpreted as the number of $d$-dimensional holes which appears at $i$ and disappear at $j$. Note that $\beta_{a_i, a_{m+1}}$ and $\beta_{a_0, a_j}$ are values out of the original filtration. In order to proceed with the calculation, we can set them as $0$. Since $a_{r+1}$ does not correspond to a simplicial complex, holes which disappear at $a_{r+1}$ may actually be considered not to disappear and persist up to infinity.
This information is summarized in the $m$-dimensional barcode (or $m$-barcode),
a multiset of intervals

\[
\left\{ \left( [b_1, d_1), c_{b_1,d_1}^{m} \right), \ldots,\left( [b_l, d_l), c_{b_l,d_l}^{m} \right) \right\},
\] 
where each interval $[b_i, d_i)$ appears $c_{b_i,d_i}^{m}$ times (its multiplicity). 
Nevertheless,
we want to use sets instead of multisets, so
a barcode, $B$, will be described as a set of intervals, each appearing repeated as many times as its multiplicity,

\[ B = \{[b_i, d_i) \}_{i=1, \ldots n} \]
We give an example in Figure \ref{fig:barcode_example}. Further details on homology and persistent homology can be found in \cite{computational}.
\subsection{Bottleneck distance and stability}\label{sec:stability}

One of the main advantages of persistent homology with respect to classical homology,
is that it is stable regarding small modifications of the input.
In order to introduce this result,
we need a notion of closeness for persistence barcodes.\\
Similarities between persistence barcodes can be  measured by bottleneck distance. 
A $\delta$-partial matching between barcodes $B_1, B_2$ is a colection of pairs $M \subset B_1 \times B_2$ such that 
\begin{itemize}
	\item For each $[b_1,d_1) \in B_1$, there is at most one $[b_2, d_2) \in B_2$, such that $\big([b_1,d_1), [b_2,d_2)\big) \in M$ and vice versa.
	\item If $\big([b_1,d_1), [b_2,d_2)\big) \in M$ then $\max\{|b_1 - d_2|, |b_1 - d_2|\} \leq \delta$
	\item If $[b,d)\in B_1$ (or $\beta\in B_2$) is unpaired, 
	then $(d-b)/2 \leq \delta$.
\end{itemize}
The bottleneck distance between two barcodes is defined as

\begin{align*}
	d_b(B_1, B_2) = \min\big\{ \delta | &\text{ there exists a }\delta\text{-matching } \\
	& \text{ between } B_1 \text{ and } B_2 \big\}
\end{align*} 
The following stability result can be found in \cite{computational}.
Given two filtration $f,g: \mathcal{K} \longrightarrow \mathbb{R}$,
we have

\[ d_b(B_f, B_g) \leq |f - g|_\infty \]
In our case,
cell tissues with similar contact network or similar centroid distribution give similar barcodes.
\subsection{Barcodes normalization}\label{sec:normalization}

Barcodes are not good for statistical analysis as shown in \cite{statistics}.
We will use numerical variables calculated from the barcodes,
but we need to deal with infinity bars before.
Consider a barcode $B$ representing a \textbf{sub} or \textbf{sup} filtration. 
We would like somehow to keep infinity bars since they gives information about the barcode.
Define the function $\xi_{z} B$ which gives the same barcode but with infinity bars $[a, \infty)$ transformed in $[a, z)$. 
In \cite{stability} it was shown that 
\[ d_b(\xi_{z} B_1, \xi_{z} B_2) \leq d_b( B_1, B_2) \]
In our sample, no cell has $15$ or more neighbors, so fixing $z=15$ will map infinity bars to bars that are always longer than the others. 
Note that \textbf{sub} and \textbf{sup} barcodes are always in the same units (number of neighboors) and can be compared between them.
In addition, 
they are invariant to rotations and scale of the input, by definition.\\
In the \textbf{rips} case,
the last complex appearing in the filtration is always contractible, so there are not infinity bars with dimensions greater than $0$,
and only one infinity interval, 
$[0, \infty)$,
in the $0$ dimensional persistent homology.
Then,
infinity bars does not give information in this context so we eliminate them using $\xi_0$. Note that this is equivalent to calculate the reduced homology.\\
Recall that we mentioned in Section~\ref{sec:rips} that the centroids from which \textbf{rips} is defined, 
still carry the units from the image and so does the barcode associated to \textbf{rips}.
Hence, it is not invariant to scale, since it depends on the distance matrix of the point cloud.
The following normalization solves
this problem: 
we are dividing each barcode by the sum of the lengths of the bars.
More specifically, given $B = \{[b_i, d_i)\}$ with no infinity bars, define
$L_B = \sum_i d_i - b_i$ and $\phi(B) = \{ [b/L_B, d/L_B ) : [b,d) \in B\}$. It is a direct consequence from \cite[lemma 3.9]{stability} that
\[ d_b(\phi(B_1), \phi(B_2)) \leq \frac{n_{max}}{\max\{ L_{B_1}, L_{B_2} \}}d_b(B_1, B_2) \]
Where $n_{max}$ is the maximum number of bars between $B_1$ and $B_2$.
Note that for any barcode $B$ coming from the 0-dimenional \textbf{rips} we have $n-1$ number of bars (one for each of the $n$ cells minus the infinity bar) and $L_B = (n-1)\bar\ell_B $, where $\bar\ell_B$ is the average length of the bars.
Then,
for all $0$-dimensional \textbf{rips} barcodes coming from our experiment

\[ d_b(\phi(B_1), \phi(B_2)) \leq  \dfrac{d_b(B_1, B_2)}{\tilde\ell} \]
where $\tilde \ell$  is the minimum of all averages $\bar \ell_B$. 
Note that $\bar \ell_B$ cannot be arbitrarily small since there are physical constraint for the size of cells in the tissue.
Unfortunately,
the $1$-dimensional case is not stable under this normalization since we cannot find a lower bound for $L$.
We drop the $1$-dimensional \textbf{rips} barcodes from the experiment.
As the following result shows, 
this normalization makes \textbf{rips} barcodes scale-invariant.
\begin{Proposition}
	Fix a normed vector space and the induced distance, $d$. 
	Fix a scalar, 
	$\alpha$, 
	and consider two point clouds $P_1$ and  $P_2 = \alpha P_1$.
	Let $B_1$ and $B_2$ be their barcodes coming from \textbf{rips}. 
	Then, 
	$\phi(B_1) = \phi(B_2)$.
\end{Proposition}
\begin{proof}
	Note that for any two points

	\[
		d(\alpha x, \alpha y) = ||\alpha x - \alpha y|| = ||\alpha(x - y)|| = |\alpha|||x - y|| = |\alpha|d(x,y)
	\]
	Then, 
	if the induced filtration by $P_1$ is $f$,
	the one from $P_2$ is $|\alpha| f$.
	In particular,
	$\beta_{a,b}^m$ in the first case must be equal to $\beta_{|\alpha| a, |\alpha| b}^m$ in the second case.
	This means that barcodes are also proportional $B_2 = |\alpha| B_1$ and $L(B_2) = |\alpha| L(B_1)$,
	so 
	\[
		\phi(B_2) = \dfrac{B_2}{L(B_2)} = \dfrac{|\alpha| B_1}{|\alpha| L(B_1)} = \phi(B_1)
	\]
\end{proof}
In particular,
it works in our setting since $\mathbb{R}^2$ with the Euclidean distance is a normed vector space.
Then, 
from each image we have five barcodes, four $\xi_{15} B$ coming from $0$ and $1$-dimensional \textbf{sub} and \textbf{sup} filtrations and $\pi \xi_0 B$ coming from $0$-dimensional \textbf{rips}.
From now on,
when we mention a barcode coming from any of these filtrations,
we assume the corresponding $\xi$ has been already applied.
\subsection{Stable topological summaries}\label{sec:variables}

In the previous section,
we saw that barcodes with the bottleneck distance are stable respect to modifications in the input.
Then,
variables defined on the barcodes,
which are stable with respect to the bottleneck distance,
will be also stable with respect to to the input.\\
In this section, we will describe the variables used in this paper and study their stability.
\subsubsection{Persistent Entropy} 

\emph{Persistent entropy} \cite{PE1,PE2} is a topological summary that can be seen as an adaptation of Shannon entropy (Shannon index in ecology) to the persistent homology context. 
Given a barcode with finite bars $B = \{[b_i,d_i)\}_{i=1 \ldots n}$ consider the length of the bars $\ell_i = d_i - b_i$  and their sum $L(B) = \ell_1 + \ldots + \ell_n$. 
Then, 
its \emph{persistent entropy} is:

\[
    PE(B) = \sum_{i=1}^n -\frac{\ell_i}{L(B)} \log \left( \frac{\ell_i}{L(B)} \right)
\]
When computed over an $m$-dimensional barcode $B$. 
The stability result appearing in \cite{stability} is simplified greatly in our case.
In particular, We have that the $0$-barcodes coming from \textbf{rips} satisfy $L(B) = 1$ after normalization.\\
First, 
recall a result relative to the Shannon entropy, 
$E_S$.
\begin{Proposition}[\mbox{\cite[p.~664]{Elements}}] \label{pro:inequality}
	Let $P$ and $Q$ be  two finite probability distributions (seen as vectors in $\mathbb{R}^{u}
	$), and let $E_S(P)$ and $E_S(Q)$ be, respectively, their  Shannon entropy. If $||P-Q||_1 \leq \frac{1}{2}$ then 

	\[
	|E_S(P)-E_S(Q)| \leq 
	||P-Q||_1\big(\log({u}
	) - \log(||P-Q||_1)\big)
	\]
\end{Proposition}
We can transform the previous proposition in the following result for persistent entropy:
\begin{Proposition}
	Let $A$ and $B$ be two barcodes with the same number of bars, $n$, all of them starting at $0$ and satisfying $L(A) = L(B) = 1$.
	If $ d_b(A,B) \leq \frac{1}{2n}$ then 

	\[
	|PE(A)-PE(B)| \leq 
	-n\,d_b(A,B)\log(d_b(A,B)).
	\]

\end{Proposition}
\begin{proof}
	Note that since both barcodes have the same number of bars and all of them start at $0$,
	the matching provided by the bottleneck distance is a one-to-one mapping between both sets of intervals.
	Then, 
	we can order the barcodes in such a way that the bars matched by bottleneck distance are listed in 
	the same position. Besides,
	since we have $L(A)=1$ we can treat its barcode
	$\{[0, d_i)\}$ as a finite probability distribution $P=\{ d_i \}$.
	Name $Q$ the probability distribution of $B$.
    Note that $||P-Q||_1 \leq n\,d_b(A,B)$.
	Then,
	substituting in the formula in Proposition~\ref{pro:inequality} 
	and using $\log(nd_b(A,B)) = log(n) + log(d_b(A,B))$,
	the result follows.
\end{proof}
Then,
this proposition gives a stability result for the \textbf{rips} filtration. In the \textbf{sub} and \textbf{sup} case the result is not straight forward,
but we can still talk about stability (see Theorem 3.12 in \cite{stability}).
Onward,
We will refer to the persistent entropy of a $d$-dimensional barcode as $PE_d(B)$.
\subsubsection{Tropical polynomials}

Tropical coordinates allow to define stable polynomials over barcodes as explained in \cite{tropical}.
These polynomials are defined on the max-plus semiring $(\mathbb{R} \cup\{-\infty\},\boxplus,\odot)$ with addition and multiplication being defined as:

\[
	a \boxplus b := \max(a,b) \qquad a \odot b := a + b.
\]
In particular,
for the variables $x_i$, 
polynomials of this semi-ring are written (with the usual notation) as

\[
	\max(a_1 + a^1_1x_1 + \ldots + a^1_q x_q, \ldots, a_r + a^r_1 x_1 + \ldots + a^r_q x_q)
,\]
where $a_i\in \mathbb{R}$ and $a_i^j\in \mathbb{N}_{0}$. If we make analogous definition for barcodes,
using the length of the bars, $\ell_i$, as variables,
polynomials of the form

\[
\max(a_1 + a^1_1 \ell_1 + \ldots + a^1_q \ell_q, \ldots, a_r + a^r_1 \ell_1 + \ldots + a^r_q \ell_q)
\]
are shown to be stable with respect to the bottleneck distance.
\begin{Proposition}\cite{tropical}
	Let $F$ be a polynomial as stated before, defined over barcodes $A,B$. 
	Then,
	there exists a constant $C$ such that

	\[ 
		|F(A) - F(B)| \leq C \,d_b (A, B)
	\]

\end{Proposition}
\subsubsection{Persistence landscapes}
\begin{figure}
	\centering
	\includegraphics[scale=1]{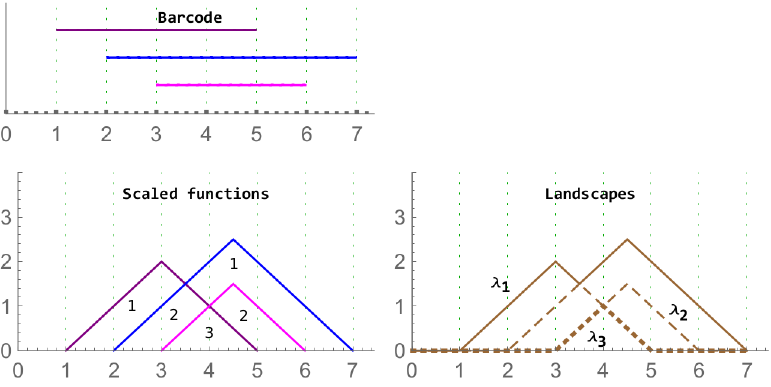}
	\caption{Left column: A barcode (top) and its corresponding rescaled rank functions (bottom). The values of the functions on the corresponding region are given. On the right, the associated landscape with the functions  $\lambda_1$, $\lambda_2$ and $\lambda_3$, displayed with different layouts.}
	\label{fig:lands}
\end{figure}

A Persistence landscape \cite{landscape} is a sequence of summary functions obtained from a barcode.
Given a barcode $B = \{ [b_i, d_i) \}$ perform the change of coordinates 

\[ 
l = \dfrac{d + b}{2} \qquad h = \dfrac{d - b}{2} 
\]
The rescaled rank function, $\lambda : \mathbb{R}^2 \rightarrow \mathbb{R}$ is defined as

\[
\lambda(l, h) = \left\{ \begin{array}{l l}
\beta_{l-h, l+h} & \text{ if } h \ge 0 \\
0 & \text{ otherwise } 
\end{array} \right.
\]
The persistence landscape is the set of function $\lambda_k: \mathbb{R} \rightarrow \mathbb{R}$ with $k\in \mathbb{N}$ given by

\[ 
\lambda_k(t) = \sup(x \ge 0\quad|\quad\beta_{t-x, t+x} \ge k) 
\]
Persistence landscape are related with tropical polynomials. 
In particular,
they are an example of what is called tropical rational function (see \cite{land-sta, tropical}). See Figure~\ref{fig:lands} for an illustration.
There is also a stability result available,
\begin{Proposition}\cite{landscape, land-sta}
Let A and B be two persistence barcodes and let $\lambda^A$ and $\lambda^B$ be their persistence landscape. Then, 
for all $k$ and $t$,

	\[
		|\lambda_k^A(t) - \lambda_k^B(t)| \le d_b(A,B) 
	\]
\end{Proposition}

Since we are interested in variables and not summary functions,
we will use the $1$-norm of $\lambda_k$.
Note that for \textbf{sub} and \textbf{sup} the domain of the landscape is restricted to $[0,15]$ and for \textbf{rips} all the intervals will lie in $[0,1]$.
Then,
in our case

\[
||\lambda_k^A - \lambda_k^B||_1 \le C\max_t(|\lambda_k^A(t) - \lambda_k^B(t)|) \le Cd_b(A,B) 
\]
where $C=15$ or $1$ depending on the filtration.\\
We have seen how to obtain numerical summaries from the tissue images,
each of them stable respect to the filtration induced by the cell organization.
This means that, if their contact network or their centroid distribution were similar (up to scaling),
the resulting variables will be similar.
We have proved these summaries satisfy the desired conditions: 
they measure topological and geometrical properties (at least up to scaling or rotation), 
they are robust to modification in the organization of the cells (we mean respect to modification in their contact network or centroid distribution) and 
all barcodes have been normalized to avoid bias when comparing them.
\section{Results}\label{sec:3}
	
	Our experiment is divided in two parts. 
	First, 
	we analyze the barcodes using statistical techniques. 
	Then, 
	we try to classify the images using Random Forests. 
	In both cases, 
	we use variables coming from the previous section.
	Before each experiment,
	we performed an exploratory analysis with many variables.
	We selected different polynomials and $k$ values for the landscapes.
	The notation is as follows: 
	$|\lambda_d^{\textbf{filt}}(k)|$ means the summary corresponding to the norm of the landscape computed from the $d$-dimensional persistence barcode of the filtration \textbf{filt} with parameter $k$.
	Instead of a fixed number,
	we may express $k$ with a percentage together with the letter $N$.
	For example, 
	$|\lambda_1^{\textbf{sub}}(0.05N)|$ means we have used the sub filtration and $k= \emph{floor}(0.0.05N)$. 
	For $N=187$ we have $k = 9$.
	$Poly_d^{\textbf{filt}}(r, k)$ means the sum $\ell_r + \ell_{r+1} + \ldots + \ell_{k}$ where $\ell_r$ is the $k$-th largest length in the $d$-dimensional barcode of the filtration \textbf{filt}. 
	Again, we may express $r$ or $k$ as percentages instead of fixed numbers.
	For example,
	$Poly_0^{\textbf{sup}}(2, 0.02N)$ means $\ell_2 + \ell_3$ in the 0-dimensional \textbf{sup} barcode when $N=187$, 
	since $k = \emph{floor}(0.02 \cdot 187) = 3$.
	If only one element appear in the sum,
	we write directly $\ell_d^{\textbf{rips}}(k)$ for the $k$-th length.
	Finally $PE_d^\textbf{filt}$ mean the persistent entropy of the $d$-dimensional \textbf{filt} barcode.
    As mentioned in Section~\ref{sec:number},
    we fix $N=187$.
	After choosing different values for the parameters of the variables in each of the $5$ barcodes (the $0$ and $1$-dimensional barcodes of \textbf{sub} and \textbf{sup}, and the $0$-dimensional barcodes of the \textbf{rips} filtration),
	we obtain a total of $57$ summary variables per image.
    The code with the whole experiment can be found in a publicly available repository \footnote{github.com/Cimagroup/topo-summaries-for-packed-tissues}.
\subsection{The statistical analysis}\label{sec:statistics}
    
    We will look for significant differences in the distributions followed by the TDA summaries in each of the tissues.
    
\subsubsection{Eplithelial tissues}

	Note that our samples are relatively small: between $12$ and $16$ images per tissue.
	Then,
	we cannot assume the variables follow any parametric distribution.
	This means that our statistical analysis must be based on a non-parametric test.
	First,
	we will use the Kruskall-Wallis test to see if each variable follows the same distribution in all the tissues.
	If this is not true,
	we will try to find differences between pair of tissues using the Dunn test.
	Since we are using many variables,
	we fix the $p$-value in $0.01$.\\
	The Kruskall-Wallis test found significant differences among the tissues for all the variables except for two. 
	Due to a high redundancy among the variables,
	results of the Dunn Test of some selected variables are shown in Table~\ref{tab:187}. 
	Note that cEE and cNT can be easily differentiated between them and from the rest using \textbf{sub} and \textbf{sup}, as expected.
	Nevertheless, 
	no differences were found between cEE vs cNT and cNT vs dNP for \textbf{rips}, 
	what means that we could not distinguish their centroid distributions. 
	Differences between dNP and both wing tissues,
	dWL and dWP,
	were found but only for \textbf{rips}.
	In particular,
	dNP vs dWL can only be differentiated by $\ell^{\textbf{rips}}_{0}(0.05N)$, $\ell^{\textbf{rips}}_{0}(0.10N)$ and $\ell^{\textbf{rips}}_{0}(0.15N)$.\\
	Finally,
	we could find differences between dWL and dWP only for one variable,
    $|\lambda_0^{\textbf{sup}}(0.03N)|$. 
    A possible explanation can be the small amount of cells selected from each image of the sample.
    With the expectation that we could find more differences by increasing the number of cells, 
    we design a more specific experiment to compare  these two tissues taking the maximum  number of available cells ($N=257$, 
    see Table~\ref{tab:number}) and performing a Mann-Whitney U test.
    In Table~\ref{tab:wings},
    the results for the test fixing $N=187$ and $N=257$ are displayed for three significant variables. 
    As we expected, 
    we find more significant differences with the increase in the number of cells. 
    Note that a change in $N$ is just a change in a parameter of the variables,
    and not a change of the sample size (because the sample images remain the same).\\
	We will analyze the meaning of some of the variables appearing in this section in Section~\ref{sec:interpretation}.
\begin{table}
\centering
	\caption{ Differences between the tissues for $187$ number of cells. A check mark implies the $p$-value of that variable is smaller than $0.01$ in the Dunn test.}\label{tab:187}
	\vspace{0.3cm}
	\centering
	\begin{tabular}{|c|c|c|c|c|c|c|}
		\hline
		\textbf{187 cells}  & $|\lambda_1^{\textbf{sub}}(0.05N)|$ &  $|\lambda_0^{\textbf{sup}}(0.03N)|$ & $\emph{Poly}_1^{\textbf{sup}}(2,0.02N)$ & $\ell^{\textbf{rips}}_{0}(0.05N)$ & $PE_0^{\textbf{rips}}$ \\ \hline
		\multicolumn{1}{|c|}{cEE vs cNT} & \checkmark  & $\times$ & $\times$ & $\times$   & $\times$   \\ \hline
		\multicolumn{1}{|c|}{cEE vs dNP} & \checkmark  & $\times$ & \checkmark & $\times$   & \checkmark   \\ \hline
		\multicolumn{1}{|c|}{cNT vs dNP} & \checkmark  & $\times$ & \checkmark & $\times$ & $\times$ \\ \hline
		\multicolumn{1}{|c|}{cEE vs dWL} & \checkmark  & $\times$ & \checkmark & \checkmark & \checkmark \\ \hline
		\multicolumn{1}{|c|}{cNT vs dWL} & $\times$    & $\times$ & \checkmark & \checkmark & \checkmark \\ \hline
		\multicolumn{1}{|c|}{dNP vs dWL} & $\times$    & $\times$   & $\times$ & \checkmark & $\times$  	 \\ \hline
		\multicolumn{1}{|c|}{cEE vs dWP} & \checkmark  & \checkmark & \checkmark & \checkmark & \checkmark \\ \hline
		\multicolumn{1}{|c|}{cNT vs dWP} & $\times$    & \checkmark & \checkmark & \checkmark & \checkmark \\ \hline
		\multicolumn{1}{|c|}{dNP vs dWP} & $\times$    & $\times$   & $\times$ & \checkmark   & \checkmark  \\ \hline
		\multicolumn{1}{|c|}{dWL vs dWP} & $\times$    & \checkmark   & $\times$ & $\times$   & $\times$   \\ \hline
	\end{tabular}
\end{table}
\begin{table}[ht]
    \caption{ The $p$-values of some variables for the Mann-Whitney U test between dWL and dWP. The number of cells is set in $187$ and $257$. }\label{tab:wings}
    \vspace{0.3cm}
    \centering
    \begin{tabular}{|c|c|c|c|c|c|}
    	\hline
    	\textbf{dWL vs dWP}  & $|\lambda_0^{\textbf{sup}}(0.03N)|$ &  $|\lambda_1^{\textbf{sup}}(0.02N)|$ & $\emph{Poly}^{\textbf{sup}}_{1}(1,0.02N)$ \\ \hline
    	\multicolumn{1}{|c|}{N = 187} & 0.013   & 0.01   & 0.019 \\ \hline
    	\multicolumn{1}{|c|}{N = 257} & 0.012      & 0.006   & 0.005  \\ \hline
    \end{tabular}
\end{table}
\subsubsection{Comparing the CVT-path with epithelia }

Some of the epithelial tissues will be compared with their most similar tessellation in the CVT-path.
Following \cite{embo},
cNT follows a similar neighbor distribution to $CVT_1$,
dWL to $CVT_4$ and dWP to $CVT_5$.
Since we are interested only in making those pairwise comparisons, 
we are performing again the Mann-Whitney U test instead of the Kruskall-Wallis one.
The minimum valid cells per image is $257$ (see Table~\ref{tab:number}).
A selection of the results are displayed in Table~\ref{tab:CVT}.
Many variables follow different distributions between the CVT tissue and its epithelium counterpart (between $8$ and $17$ depending on the type compared), 
most of them in the \textbf{rips} filtration.
Differences in the \textbf{sub} and \textbf{sup} filtrations were only found between cNT and $CVT_1$.

\begin{table}[ht]
	\caption{ Differences between some tissues and their CVT path counterpart. A check mark implies the $p$-value of that variable is smaller than $0.01$ in the Mann-Whitney U test.}\label{tab:CVT}
	\vspace{0.3cm}
	\centering
	\begin{tabular}{|c|c|c|c|c|c|}
		\hline
		\textbf{257 cells}  & $|\lambda_0^{\textbf{sub}}(0.15N)|$ &  $PE_0^{\textbf{rips}}$ & $\emph{Poly}^{\textbf{rips}}_{0}(1,0.05N)$ & $\ell^{\textbf{rips}}_{0}(0.10N)$ \\ \hline
		\multicolumn{1}{|c|}{cNT vs $CVT_1$} & \checkmark    & \checkmark   & \checkmark & \checkmark  \\ \hline
		\multicolumn{1}{|c|}{dWL vs $CVT_4$} & $\times$      & \checkmark   & \checkmark & $\times$    \\ \hline
		\multicolumn{1}{|c|}{dWP vs $CVT_5$} & $\times$      & \checkmark   & $\times$   & \checkmark  \\ \hline
	\end{tabular}
\end{table}
\subsection{Classifying the images}

We will classify the epithelial images in three classes: cEE, cNT and Drosophila tissues.
Drosophila tissues are dWL, dWP and dNP. 
These tissues can be easily separated from $cEE$ and $cNT$ using the mean and variance of the degrees in the contact graph.
Nevertheless,
distinguishing between cEE and cNT is more difficult.
Since we do not have a big sample of data,
we will use the Random Forests technique to avoid over-fitting.
Many variables used in network analysis have a strong relation with the mean degree in this specific context \cite{Villoutreix}[Sec. 8.2.2].
Then,
variables used for the network analysis are:
the mean and variance of the degree and the amount of cells with degree equals to $2,\,3,\,4, \ldots 13$ cells. 
We fix $N=187$ as in the first experiment of Section~\ref{sec:statistics}.
We use $3/4$ of data as a training test and $1/4$ for validation. 
We fix the number of trees in $200$ since the accuracy is already stabilized for that number.
This procedure is repeated $10^4$ times,
the average accuracy of the classification is shown in Table~\ref{tab:classification}.
The best results is reached with only $3$ variables: the mean and variance of the degree and $|\lambda_0^{\textbf{sub}}(0.10N)|$. 
The validation results are slightly better than the training ones,
so we do not commit over-fitting.
The selected variables outperform the others.
This proves that TDA variable may be useful to complement other variables in machine learning tasks.

\begin{table}[ht]
	\caption{ Classification of tissues using all TDA variables, all network variables, all variables together, mean and variance of the degree and a combination of these two with $|\lambda_0^{\textbf{sub}}(0.10N)|$. The accuracy for the training/validation sets are displayed. Best results are highlighted.}\label{tab:classification}
	\vspace{0.3cm}
	\centering
	\begin{tabular}{|c|c|c|c|c|c|} 
		\hline
		\textbf{187 cells} & TDA       & network          & mixed            & m \& v            & \multicolumn{1}{c|}{\begin{tabular}[c]{@{}c@{}}m \& v \&\\ $|\lambda_0^{\textbf{sub}}(0.10N)|$ \end{tabular}}  \\ 
		\hline
		cEE                & 85.5/86.7 & 86.2/87.8         & 86.1/87.8       & 89.7/94.3          & \textbf{97.1/98.7}                                                                                                                 \\ 
		\hline
		cNT                & 82.6/83 & 93.4/93.7 & 92.2/92.9          & 89.2/89.7       & \textbf{96.2/97.7}                                                                                                                        \\ 
		\hline
		Drosophila         & 99.7/99.8   & \textbf{100/100} & \textbf{100/100} & \textbf{100/100} & \textbf{100/100}                                                                                                                 \\ 
		\hline
		overall            & 92.9/93.4   & 95.8/96.1          & 95.5/96          & 95.5/96.2          & \textbf{98.6/99}                                                                                                                 \\
		\hline
	\end{tabular}
\end{table}
\subsection{Interpretation of the variables}\label{sec:interpretation}

As we will see,
information carried by \textbf{sub} and \textbf{sup} filtration are strongly related with the neighbor distribution,
but not only, since details of the inner organization of the tissue might enrich it.
Besides,
\textbf{rips} is strongly related with the relative proximity of the centroids of the cells inside the same image and any interpretation of a variable must be in that terms.
In the sequel,
we will use the term $n$-cell to refer to a cell with $n$ neighbors.
\subsubsection{The variable $|\lambda_1^{\textbf{sub}}(k)|$}

\begin{figure}
	\centering
	\includegraphics[scale=0.6]{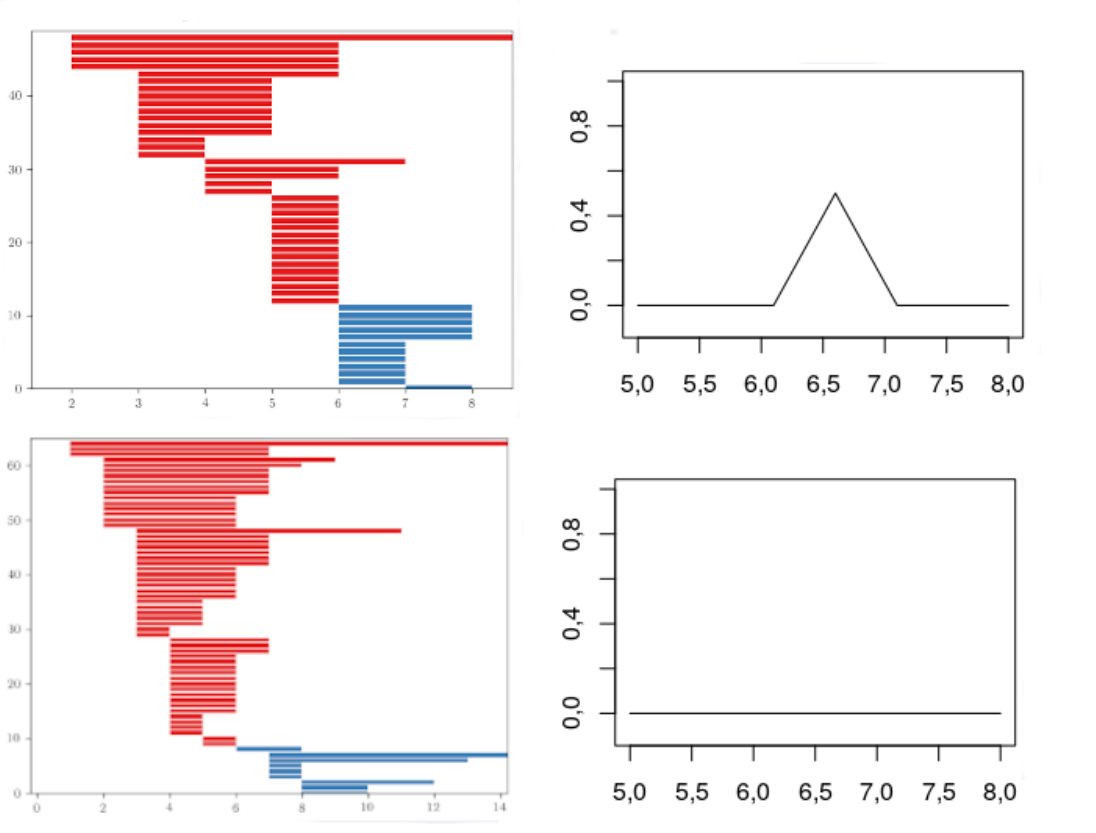}
	\caption{An example of barcodes ($0$-bars in red and $1$-bars in blue) with their landscape $\lambda_1^{\textbf{sub}}(9)$. The top corresponds to dNP and the bottom to cEE. Note that in the cEE case there are not $9$ $1$-dimensional holes simultaneously alive, so its landscape is zero.}
	\label{fig:lambda_1_sub_barcodes}
\end{figure}

This landscape is detecting when there are at least $m$ $1$-dimensional holes alive.
For $N=187$ we obtained the most discriminating $m$ value is
$\emph{floor}(0.0.05N) = 9$,
see Figure~\ref{fig:lambda_1_sub_barcodes}.
$1$-dimensional holes in the \textbf{sub} filtration are formed when there are clusters of cells surrounded by other cells with a smaller amount of neighbors, see  Figure~\ref{fig:filt_sub_dnp_cee}.
For example,
cEE has a big variance with more cells with few neighbors ($2$ or $3$) or many neighbors ($8$ or $9$) than other tissues.
In particular, 
cells with more neighbors have a greater chance to appear forming clusters than in other tissues,
where is more common to find them isolated.
Then,
a smaller number of $1$-dimensional holes is expected.
There is another factor,
cEE cells are far away of being convex allowing
settings like isolated small cells embedded between two or three big cells.
So again,
$1$-dimensional holes in cEE are less likely that in other tissues,
and never reach the $9$ simultaneous $1$-dimensional holes threshold.\\
On the other hand,
Drosophila images have small variance with plenty of cells with $6$ neighbors.
In particular,
when $i=6$  all  $6$-cells are connected and many $1$-dimensional holes appear,
one for each cluster formed by cells with more than $6$ neighbors.
See Figure~\ref{fig:filt_sub_dnp_cee}.\\
Finally,
cNT is halfway of both.\\
In general,
there is a strong correlation between this variable and the variance of the degrees.
For the rest of variables, complementary information will become more important than just the degree distributions.
\subsubsection{The variable $|\lambda_0^{\textbf{sub}}(k)|$}

\begin{figure}
	\centering
	\includegraphics[width=\textwidth]{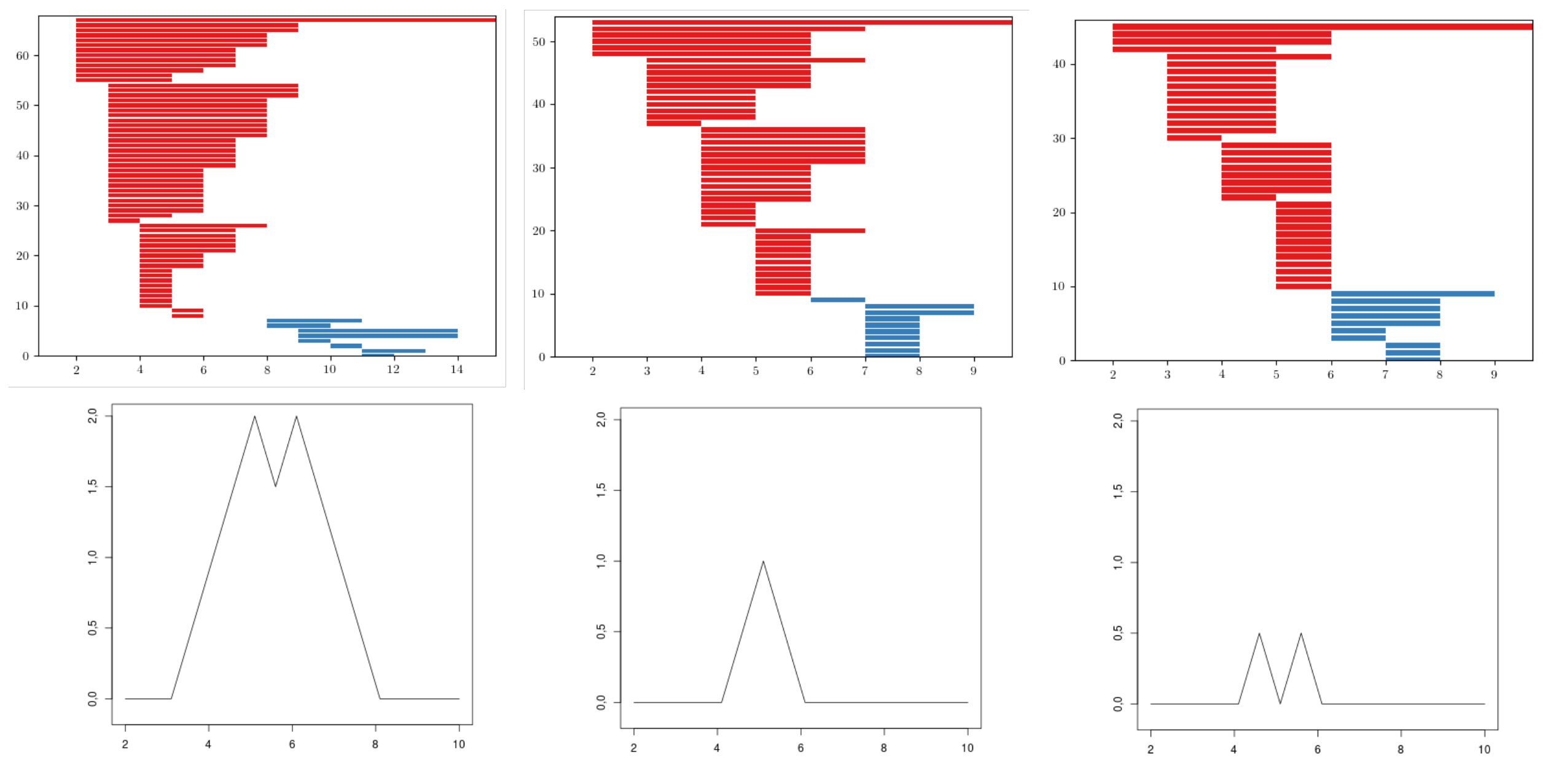}
	\caption{An example of barcodes and its landscapes $\lambda_0^{\textbf{sub}}(18)$ from cEE, cNT and dWL respecively. Note that the domain of the landscapes of cNT and dWL are the same, but the area of cNT is greater, since in that domain the $18$ bars are the same while for dWL the bars vary.}
	\label{fig:lambda_0_sub_barcodes}
\end{figure}

In this case,
landscape is detecting when there are at least $m$ connected components simultaneously alive. 
In our experiment, 
the best $m$ is  $\emph{floor}(0.0.10N) = 18$.
An interesting pattern arise, which improves the results with respect to just using the number of neighbors.
Many cells with $2$ or $3$ neighbors are just cells on the boundary which will be connected soon with some neighbor.
Nevertheless, cEE tissues,
have non-boundary cells with $2$ or $3$ neighbors which are isolated and surrounded by cells with $6$ or $7$ neighbors. 
That will generate some longest connected components than in the other tissues.\\
In the other chicken tissue,
cNT, 
cells with degree from $4$ to $8$ neighbors are more uniformly distributed in the image. 
Hence, 
a greater proportion of connected components arises with birth time $4$ or $5$ and death time for $6$ neighbors.\\
This effect is even clearer in Drosophila tissues: since there are fewer $4$-cells,
connected components with $2$ or $3$ cells on the boundary are alive until cells with $5$ neighbors appear.
Many of them connect with cells on the boundary,
but other $5$-cells keep isolated or in small clusters creating connected components.
These new connected components have a short life and usually die when $6$-cells appear.\\
Then,
cEE landscape tends to have a greater area with two close peaks,
cNT landscape tends to have a medium area with only one peak and Drosophila landscape tends to have a smaller area and $1$ or $2$ separated peaks depending if there have enough $2,3$-cells or not;
see Figure~\ref{fig:lambda_0_sub_barcodes}.\\
Therefore, 
this variable is not only taking into account the distribution of the cells but how cells with different number of neighbors are connected among them and with regard to the boundary.
As it is shown in the Random Forests classification,
this variable performs better for classification than comparing directly the number of neighbors (see the accuracy of network variables vs mean + variance + $|\lambda_0^{\textbf{sub}}(0.10N)|$ in Table~\ref{tab:classification}).
\subsubsection{$|\lambda_1^{\textbf{sup}}(k)|$ and $\emph{Poly}^{\textbf{sup}}_{1}(1,k)$}

\begin{figure}
	\centering
	\includegraphics[width=\textwidth]{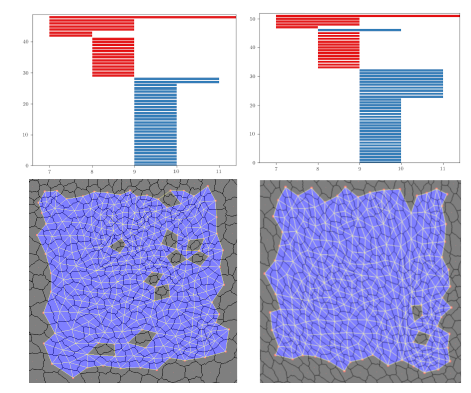}
	\caption{An example with the persistence barcodes of dWL and dWP which give the median for $|\lambda_1^{\textbf{sup}}(k)|$ and their corresponding \textbf{sup} filtration when i=10.}
	\label{fig:dwlvsdwp}
\end{figure}

These variables become specially important when comparing dWL and dWP.
Actually,
in this case there is a huge correlation between both variables since they are measuring the same feature at the tissue level.
1-dimensional holes in the \textbf{sup} filtrations appear when there are cells (or cluster of cells) with a small number of neighbors which are surrounded by cells with a higher number of neighbors.
In this case,
the key difference between both tissues is provided by persistence bars which appear when there are 4-cells, 
some of whose neighbors are 6 or 7-cells.
The presence of this combination of cells provides more bars with longer persistence in dWL than in dWP,
see Figure~\ref{fig:dwlvsdwp}.
Then, 
$|\lambda_1^{\textbf{sup}}(k)|$ (and the sum $\emph{Poly}^{\textbf{sup}}_{1}(1,k)$) will be greater in dWL.
For $N=257$,
the best result is reached when $k = \emph{floor}(0.0.02N) = 3$.
In Table~\ref{tab:wings_sup},
a small experiment showing this variable is measuring different features than just the neighbors distribution is displayed.
\begin{table}[ht]
	\caption{Classification using Random Forests of the epithelial tissues. On the left, results using the number of neigbors and mean and variance of the degree. It can be seen that $|\lambda_1^{\textbf{sup}}(3)|$ performs better}\label{tab:wings_sup}
	\vspace{0.3cm}
	\centering
	\begin{tabular}{|c|c|c|}
		\hline
		\textbf{257 cells}  & Network &  $|\lambda_1^{\textbf{sup}}(3)|$ \\ \hline
		\multicolumn{1}{|c|}{dWL} & \textbf{63.8/67.4} &  59.9/59.9  \\ \hline
		\multicolumn{1}{|c|}{dWP} & 64.9/66.5 &  \textbf{82.6/82.8} \\ \hline
		\multicolumn{1}{|c|}{global} & 63.1/65.5 &  \textbf{70/71.8} \\ \hline
	\end{tabular}
\end{table}
\subsubsection{$\ell_0^{\textbf{rips}}(k)$}

\begin{figure}
	\centering
	\includegraphics[width=\textwidth]{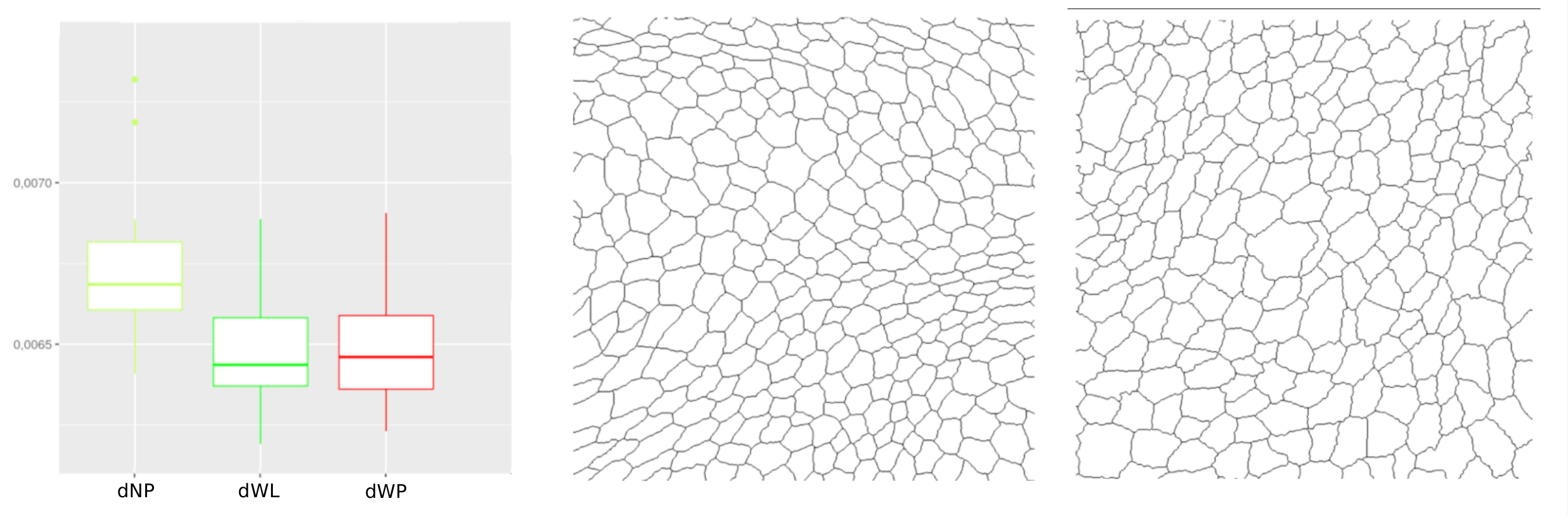}
	\caption{On the left, the bloxpot corresponding to $\ell_0^{\textbf{rips}}(0.10N)$. On the right, the images of dNP and dWL which provide the median.}
	\label{fig:length_0_rips}
\end{figure}

Calculate the length of the $m$-th longest bar in the Rips filtration is equivalent to the distance for which there are less than $m$ connected components.
Since our barcodes are normalized,
the distance have no units and can be interpreted as a proportion.
Besides,
in this data set the longest bars are associated to connected components corresponding to isolated centroids, or small clusters of centroids (recall the infinity bar was removed).
Then,
this variable is directly related with the (relative) distance between the centroids.
In practice,
it is measuring if there are at least $m$ centroids with a relative distance to the main connected component bigger than the others in the same image.
The most discriminating $m$ for $N=187$ is $\emph{floor}(0.0.10N) = 18$.
This type of variable becomes important when analyzing Drosophila tissues,
since it is the only one finding differences between dNP and the others (dWL, dWP).
An example is shown in Figure~\ref{fig:length_0_rips}.
\subsubsection{$PE_0^{\textbf{rips}}$}

Note that since Shannon entropy is a concave function in the space of probability distribution \cite{Elements}, so is the persistent entropy in the space of normalized \textbf{rips} barcodes.
Then,
knowing the maximum will give us a valuable hint for understanding this variable at the tissue level.\\
We will consider the point cloud as a finite metric space, $M$. 
Then, 
we define a distance graph on $M$, $G_M$,  as a clique graph whose set of vertices is given by the centroids and the weight on each edge is the distance between the corresponding vertices.
\begin{Proposition}
Let $M$ be a finite metric space and $B_M$ its persistence barcode (with the infinity bar removed) coming from the \textbf{rips} filtration over $G_M$. 
Then, 
$PE_0(B_M)$ is maximum if and only if the minimum spanning tree of its distance graph has a constant weight for all its edges.
\end{Proposition}
\begin{proof}
    Note that since we only consider the $0$-dimensional \textbf{rips} barcode,
    all bars are born at $0$and their death value is the same than their length.
    Besides,
    note that Shannon entropy reaches its maximum value when all probabilities are the same \cite{Elements}. 
    Then,
    persistent entropy reaches its maximum value when all bars have the same length.
    Combining these two facts,
    we have that persistent entropy will be maximum if and only if all bars have the same death value.
    One direction is clear:
    If the minimum spanning tree has a constant weight,
    it means that all the vertices are isolated until the filtration value reaches that constant.
    Then, 
    all the finite bars die at that value.
    For proving the other direction,
    assume the minimum spanning tree does not have a constant weight and define $w$ as its minimum weight.
    When the filtration value is $w$,
    some of the vertices become connected between them but not all of them.
    This means that some bars die at $w$, but some will die later.
    Then,
    persistent entropy cannot be maximum.
\end{proof}
Then,
$PE_0^{\bf rips}$ is strongly related with the variability of the weights in the minimum spanning tree of the distance graph.
It makes sense since entropy may also be understood as a diversity index.
In particular,
it may have an interpretation in terms of how centroids of the cells are related between them.
This variable becomes specially important when comparing some tissues with their CVT counterpart,
since it was the only one (together with $\emph{Poly}_0^{\bf rips}(1,0.05N)$) succeeding in differentiating all the cases, see Table~\ref{tab:CVT}. 
Then, 
it means that the CVT-path are failing to imitate the centroid distribution of the cells.
\section{Discussion}\label{sec:4}

We summarize here the results of this paper.
Normalizing the number of cells obtained from each tissue,
as well as the \textbf{rips} barcodes, 
allowed us to compare the network and centroid distributions of different cell tissues without loosing stability properties.
This was proven by some theoretical results appearing in Sections~\ref{sec:normalization} and \ref{sec:variables}.
Note that these results may be generalized with respect to other tessellations of the plane.
In Section \ref{sec:3}, 
we compared some epithelia, obtaining some conclusions that might be of  interest for the biological community:
\begin{itemize}
    \item The geometry of the cells in cEE and cNT are completely different (cNT cells tend to be convex, while cEE do not). 
    Nevertheless,
    their centroid distributions turned out to be very similar.
    Is there any biological or physical reason for this fact?
    \item Wing tissues in different state of development, like dWL and dWP, are difficult to differentiate. 
    However, the variable $\lambda_1^\textbf{sup}(k)$ and $\emph{Poly}_1^{\textbf{sup}}(1,k)$ worked pretty well in this context.
    The main reason was a difference in the number of $4$-cells surrounded by a mix of $6$ and $7$-cells.
    Is this kind of cell combination a good indicator for the level of development?
    If so,
    why?
\end{itemize}
There are also interesting results for the pattern recognition community:
\begin{itemize}
    \item We provided an example where TDA may be useful to study networks with a very simple topology,
    leading to the study of variables which would have been difficult to discover otherwise.
    \item In particular,
    we propose a combination of normalization in the original image and in the barcode which allows to prove formal stability of the method.
    \item This paper also provides an example where TDA variables may be combined with others to improve machine learning performance.
\end{itemize}
We have also seen that, 
despite the fact that CVT-path is good imitating the polygonal distribution of some of the tissues,
they may fail to model their centroid distribution. 
A future work could be to use \textbf{rips} variables,
specially persistent entropy,
to improve the simulation of epithelium.
Another interesting future work is to adapt this analysis to 3D epithelium,
as well as to other fields,
such as material science.

\vspace{15pt} 


\textbf{Acknowledgments:} Authors would like to thanks researchers L.M. Escudero, P. Gómez-Gálvez and C. Molero-Ríos for their valuable help during the development of this research. \\

\textbf{Funding:} This research was funded by Ministerio de Ciencia e Innovación – Agencia Estatal de Investigación /10.13039/501100011033, grant number PID2019-107339GB-I00. The author M. Soriano-Trigueros was partially funded by the grant VI-PPITUS from University of Seville.


\bibliographystyle{ieeetr}
\bibliography{mybibfile.bib}

\begin{thebibliography}{10}

\bibitem{cell1}
V.~Emmanuele, A.~Kubota, and M.~H. et~al., ``Fhl1 w122s causes loss of protein
  function and late-onset mild myopathy,'' {\em Hum. Mol. Genet.}, vol.~24,
  no.~3, pp.~714--726, 2014.

\bibitem{cell2}
J.-A. Park, J.~H. Kim, and J.~F. et~al., ``Unjamming and cell shape in the
  asthmatic airway~epithelium,'' {\em Nat. Mater.}, vol.~14, no.~10,
  pp.~1040--1048, 2015.

\bibitem{cell3}
D.~S{\'{a}}nchez-Guti{\'{e}}rrez, A.~S{\'{a}}ez, A.~Pascual, and L.~Escudero,
  ``Topological progression in proliferating epithelia is driven by a unique
  variation in polygon distribution,'' {\em {PLoS} {ONE}}, vol.~8, no.~11,
  p.~e79227, 2013.

\bibitem{convergencia}
M.~Emelianenko, L.~Ju, and A.~Rand, ``Nondegeneracy and weak global convergence
  of the lloyd algorithm in $\mathbb{R}^d$,'' {\em SIAM Journal on Numerical
  Analysis}, vol.~46, no.~3, pp.~1423--1441, 2008.

\bibitem{embo}
D.~Sanchez-Gutierrez, M.~Tozluoglu, and L.~M.~E. et~al., ``Fundamental physical
  cellular constraints drive self-organization of tissues,'' {\em The {EMBO}
  J.}, vol.~35, no.~1, pp.~77--88, 2016.

\bibitem{epigraph}
P.~Vicente-Munuera, P.~Gomez-Galvez, R.~J. Tetley, C.~Forja, A.~Tagua,
  M.~Letran, M.~Tozluoglu, Y.~Mao, and L.~M. Escudero, ``Epigraph: an
  open-source platform to quantify epithelial organization,'' {\em
  Bioinformatics}, vol.~36, no.~4, pp.~1314--1316, 2019.

\bibitem{Villoutreix}
P.~Villoutreix, {\em Randomness and variability in animal embryogenesis, a
  multi-scale approach}.
\newblock PhD thesis, Université Sorbonne Paris Cité, 2015.

\bibitem{ELZ00}
H.~Edelsbrunner, D.~Letscher, and A.~Zomorodian, ``Topological persistence and
  simplification,'' {\em Discret. {\&} Comput. Geom.}, vol.~28, no.~4,
  pp.~511--533, 2002.

\bibitem{ZC05}
A.~Zomorodian and G.~Carlsson, ``Computing persistent homology,'' {\em Discret.
  {\&} Comput. Geom.}, vol.~33, no.~2, pp.~249--274, 2004.

\bibitem{tumor}
T.~Qaiser, Y.-W. Tsang, and N.~R. et~al., ``Fast and accurate tumor
  segmentation of histology images using persistent homology and deep
  convolutional features,'' {\em M\'ed Image Anal.}, vol.~55, pp.~1--14, 2019.

\bibitem{inmuno}
E.~Merelli, M.~Rucco, P.~Sloot, and L.~Tesei, ``Topological characterization of
  complex systems: Using persistent entropy,'' {\em Entropy}, vol.~17, no.~12,
  pp.~6872--6892, 2015.

\bibitem{lung}
F.~Belchi, M.~Pirashvili, and J.~B. et~al., ``Lung topology characteristics in
  patients with chronic obstructive pulmonary disease,'' {\em Sci. Rep.},
  vol.~8, no.~1, 2018.

\bibitem{tropical}
S.~Kališnik, ``Tropical coordinates on the space of persistence barcodes,''
  {\em Foundations of Computational Mathematics}, no.~19, pp.~101--129, 2018.

\bibitem{IWCIA17}
M.~J. Jimenez, M.~Rucco, P.~Vicente-Munuera, P.~G{\'{o}}mez-G{\'{a}}lvez, and
  L.~M. Escudero, ``Topological data analysis for self-organization of
  biological tissues,'' in {\em Lect. Notes in Comput. Sci.}, pp.~229--242,
  Springer International Publishing, 2017.

\bibitem{stability}
N.~Atienza, R.~Gonz{\'{a}}lez{-}D{\'{\i}}az, and M.~Soriano{-}Trigueros, ``On
  the stability of persistent entropy and new summary functions for topological
  data analysis,'' {\em Pattern Recognition}, vol.~107, p.~107509, 2020.

\bibitem{ctic}
N.~Atienza, L.~M. Escudero, M.~J. Jimenez, and M.~Soriano-Trigueros,
  ``Characterising epithelial tissues using persistent entropy,'' in {\em
  Computational Topology in Image Context}, pp.~179--190, Springer
  International Publishing, Dec. 2018.

\bibitem{computational}
H.~Edelsbrunner and J.~Harer, {\em Computational topology : an introduction}.
\newblock American Mathematical Society, 2010.

\bibitem{collective}
D.~Bhaskar, W.~Y. Zhang, and I.~Y. Wong, ``Topological data analysis of
  collective and individual epithelial cells using persistent homology of
  loops,'' 2021.

\bibitem{images}
L.~Escudero, L.~da~Costa, and M.~M.~B. et~al., ``Epithelial organisation
  revealed by a network of cellular contacts,'' {\em Nat. Commun.}, vol.~2, sep
  2011.

\bibitem{KJRS2016}
S.~Kaliman, C.~Jayachandran, F.~Rehfeldt, and A.-S. Smith, ``Limits of
  applicability of the voronoi tessellation determined by centers of cell
  nuclei to epithelium morphology.,'' {\em Front. Physiol.}, vol.~7, no.~551,
  2016.

\bibitem{statistics}
Y.~Mileyko, S.~Mukherjee, and J.~Harer, ``Probability measures on the space of
  persistence diagrams,'' {\em Inverse Probl.}, vol.~27, p.~124007, nov 2011.

\bibitem{PE1}
H.~Chintakunta, T.~Gentimis, R.~Gonzalez-Diaz, M.~Jimenez, and H.~Krim, ``An
  entropy-based persistence barcode,'' {\em Pattern Recognit.}, vol.~48,
  pp.~391--401, Feb. 2015.

\bibitem{PE2}
M.~Rucco, F.~Castiglione, E.~Merelli, and M.~Pettini, ``Characterisation of the
  idiotypic immune network through persistent entropy,'' in {\em Proceedings of
  {ECCS} 2014}, pp.~117--128, Springer International Publishing, 2016.

\bibitem{Elements}
T.~Cover and J.~Thomas, {\em Elements of Information Theory}.
\newblock Wiley Series in Telecommunications and Signal Processing, 2nd~ed.,
  2006.

\bibitem{landscape}
P.~Bubenik, ``Statistical topological data analysis using persistence
  landscapes,'' {\em J. Mach. Learn. Res.}, vol.~16, pp.~77--102, Jan. 2015.

\bibitem{land-sta}
P.~Bubenik, ``The persistence landscape and some of its properties,'' {\em
  Topological Data Analysis. Abel Symposia}, vol.~15, pp.~97--117, 2020.

\end{thebibliography}

\end{document}